\documentclass[11pt,a4paper]{article}
\usepackage[hyperref]{acl2018}
\usepackage{times}
\usepackage{latexsym}
\usepackage{graphicx}
\graphicspath{ {} }
\usepackage{xcolor}
\usepackage{amsmath,amsfonts,amsthm,bm} 
\usepackage{subcaption}

\usepackage{url}

\aclfinalcopy 

\title{Structured Fusion Networks for Dialog}
\author{Shikib Mehri\thanks{*~Equal contribution.}~, Tejas Srinivasan$^*$, and Maxine Eskenazi \\
  Language Technologies Institute, Carnegie Mellon University \\
  {\tt \{amehri,tsriniva,max+\}@cs.cmu.edu}}

\date{}

\begin{document}
\maketitle
\begin{abstract}
Neural dialog models have exhibited strong performance, however their end-to-end nature lacks a representation of the explicit structure of dialog. This results in a loss of generalizability, controllability and a data-hungry nature. Conversely, more traditional dialog systems do have strong models of explicit structure. This paper introduces several approaches for explicitly incorporating structure into neural models of dialog. Structured Fusion Networks first learn neural dialog modules corresponding to the structured components of traditional dialog systems and then incorporate these modules in a higher-level generative model. Structured Fusion Networks obtain strong results on the MultiWOZ dataset, both with and without reinforcement learning. Structured Fusion Networks are shown to have several valuable properties, including better domain generalizability, improved performance in reduced data scenarios and robustness to divergence during reinforcement learning.
\end{abstract}

\section{Introduction}

\begin{figure}[t]

    \includegraphics[width=1\linewidth]{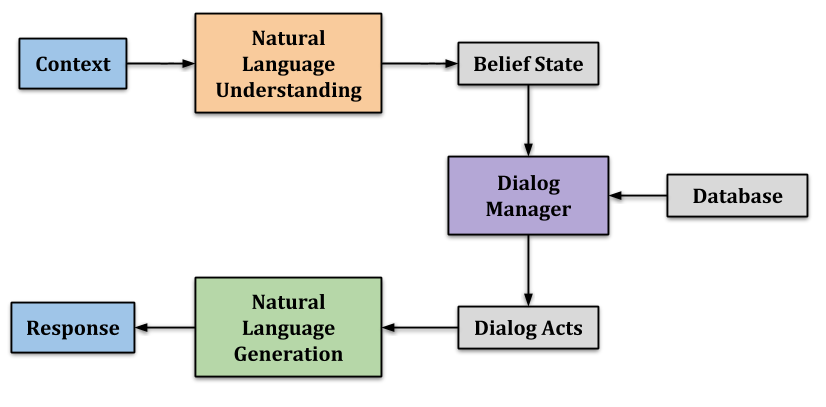}
    \caption{A traditional dialog system consisting of a natural language understanding (NLU), dialog manager (DM) and natural language generation (NLG).}
    \label{traditional}
    \vspace{-1.0em}
\end{figure}

End-to-end neural dialog systems have shown strong performance \citep{vinyals2015neural,dinan2019second}. However such models suffer from a variety of shortcomings, including: a data-hungry nature \citep{zhao2018zero}, a tendency to produce generic responses \citep{li2016deep}, an inability to generalize \citep{mo2018cross,zhao2018zero}, a lack of controllability \citep{hu2017toward}, and divergent behavior when tuned with reinforcement learning \citep{lewis2017deal}. Traditional dialog systems, which are generally free of these problems, consist of three distinct components: the natural language understanding (NLU), which produces a structured representation of an input (e.g., a belief state); the natural language generation (NLG), which produces output in natural language conditioned on an internal state (e.g. dialog acts); and the dialog manager (DM) \citep{bohus2009ravenclaw}, which describes a policy that combines an input representation (e.g., a belief state) and information from some database to determine the desired continuation of the dialog (e.g., dialog acts). A traditional dialog system, consisting of an NLU, DM and NLG, is pictured in Figure \ref{traditional}.

The structured components of traditional dialog systems facilitate effective generalizability, interpretability, and controllability. The structured output of each component allows for straightforward modification, understanding and tuning of the system. On the other hand, end-to-end neural models of dialog lack an explicit structure and are treated as a black box. To this end, we explore several methods of incorporating the structure of traditional dialog systems into neural dialog models.

First, several neural \textit{dialog modules} are constructed to serve the role of the NLU, the DM and the NLG. Next, a number of methods are proposed for incorporating these dialog modules into end-to-end dialog systems, including Na\"ive Fusion, Multitask Fusion and Structured Fusion Networks (SFNs). This paper will show that SFNs obtain strong results on the MultiWOZ dataset \citep{budzianowski2018MultiWOZ} both with and without the use of reinforcement learning. Due to the explicit structure of the model, SFNs are shown to exhibit several valuable properties including improved performance in reduced data scenarios, better domain generalizability and robustness to divergence during reinforcement learning \citep{lewis2017deal}. 

\section{Related Work}

\subsection{Generation Methods}

\citet{vinyals2015neural} used a sequence-to-sequence network \citep{sutskever2014sequence} for dialog by encoding the conversational context and subsequently generating the reply. They trained and evaluated their model on the OpenSubtitles dataset \citep{tiedemann2009news}, which contains conversations from movies, with a total of $62$M training sentences. 

Most research on generative models of dialog has built on the baseline introduced by \citet{vinyals2015neural} by incorporating various forms of inductive bias \citep{mitchell1980need} into their models, whether it be through the training procedure, the data or through the model architecture. \citet{li2015diversity} use Maximum Mutual Information (MMI) as the objective function, as a way of encouraging informative agent responses. \citet{serban2016building} proposes to better capture the semantics of dialog with the use of a hierarchical encoder decoder (HRED), comprised of an utterance encoder, a conversational context encoder, and a decoder. \citet{li2016deep} incorporate a number of heuristics into the reward function, to encourage useful conversational properties such as informativity, coherence and forward-looking. \citet{li2016persona} encodes a speaker's persona as a distributed embedding and uses it to improve dialog generation. \citet{liu2016joint} simultaneously learn intent modelling, slot filling and language modelling. \citet{zhao2017generative} enables task-oriented systems to make slot-value-independent decisions and improves out-of-domain recovery through the use of entity indexing and delexicalization. \citet{wu2017end} present Recurrent Entity Networks which use action templates and reasons about abstract entities in an end-to-end manner. \citet{zhao2018zero} present the Action Matching algorithm, which maps utterances to a cross-domain embedding space to improve zero-shot generalizability. \citet{mehri2019pretraining} explore several dialog specific pre-training objectives that improve performance on dowstrean dialog tasks, including generation. \citet{chen2019semantically} present a hierarchical self-attention network, conditioned on graph structured dialog acts and pre-trained with BERT \citep{devlin2018bert}.





\subsection{Generation Problems}
Despite their relative success, end-to-end neural dialog systems have been shown to suffer from a number of shortcomings. \cite{li2016deep} introduced the dull response problem, which describes how neural dialog systems tend to produce generic and uninformative responses (e.g., \textit{"I don't know"}). \citet{zhao2018zero} describe generative dialog models as being data-hungry, and difficult to train in low-resource environments. \citet{mo2018cross,zhao2018zero} both demonstrate that dialog systems have difficulty generalizing to new domains. 
\citet{hu2017toward} work on the problem of controllable text generation, which is difficult in sequence-to-sequence architectures, including generative models of dialog. 

\citet{wang2016diverse} describe the problem of the \textit{overwhelming implicit language model} in image captioning model decoders. They state that the decoder learns a language generation model along with a policy, however, during the process of captioning certain inputs, the decoder's implicit language model overwhelms the policy and, as such, generates a specific output regardless of the input (e.g., if it generates 'giraffe', it may always output 'a giraffe standing in a field', regardless of the image). In dialog modelling, this problem is observed in the output of dialog models fine-tuned with reinforcement learning \citep{lewis2017deal,zhao2019rethinking}. Using reinforcement learning to fine-tune a decoder, will likely place a strong emphasis on improving the decoder's policy and un-learn the implicit language model of the decoder. To this end, \citet{zhao2019rethinking} proposes Latent Action Reinforcement Learning which does not update the decoder during reinforcement learning. 

The methods proposed in this paper aim to mitigate these issues by explicitly modelling structure. Particularly interesting is that the structured models will reduce the effect of the \textit{overwhelming implicit language model} by explicitly modelling the NLG (i.e., a conditioned language model). This should lessen the divergent effect of reinforcement learning \citep{lewis2017deal,zhao2019rethinking}.



\subsection{Fusion Methods}

This paper aims to incorporate several pre-trained \textit{dialog modules} into a neural dialog model. A closely related branch of research is the work done on fusion methods, which attempts to integrate pre-trained language models into sequence-to-sequence networks. Integrating language models in this manner is a form of incorporating structure into neural architectures. The simplest such method, commonly referred to as \textbf{Shallow Fusion}, is to add a language modeling term, $p_{LM}(y)$, to the cost function during inference \citep{chorowski2016towards}. 

To improve on this, \citet{gulcehre2015using} proposed \textbf{Deep Fusion}, which combines the states of a pre-trained machine translation model’s decoder and a pre-trained language model by concatenating them using a gating mechanism with trained parameters. The gating mechanism allows us to decide how important the language model and decoder states are at each time step in the inference process. However, one major drawback of Deep Fusion is that the sequence-to-sequence model is trained independently from the language model, and has to learn an implicit language model from the training data.

\textbf{Cold Fusion} \citep{sriram2017cold} deals with this problem by training the sequence-to-sequence model along with the gating mechanism, thus making the model aware of the pre-trained language model throughout the training process. The decoder does not need to learn a language model from scratch, and can thus learn more task-specific language characteristics which are not captured by the pre-trained language model (which has been trained on a much larger, domain-agnostic corpus). 

\section{Methods}

This section describes the methods employed in the task of dialog response generation. In addition to the baseline model proposed by \citet{budzianowski2018MultiWOZ}, several methods of incorporating structure into end-to-end neural dialog models are explored. 

\subsection{Sequence-to-Sequence}

\begin{figure}[h]

    \includegraphics[width=1\linewidth]{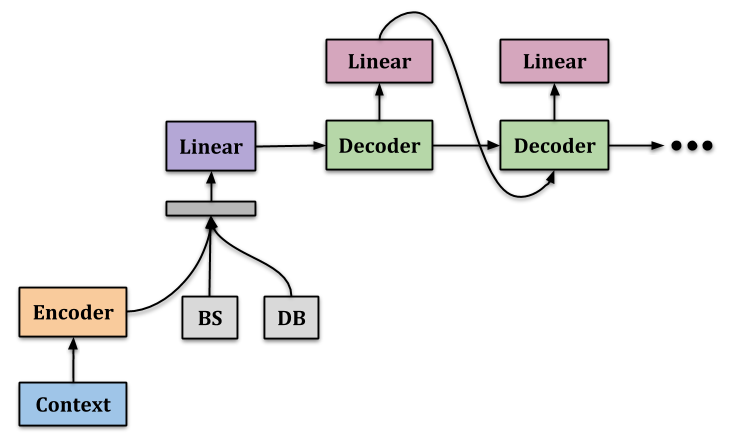}
    \caption{A diagram of the baseline sequence-to-sequence architecture. The attention mechanism is not visualized, however experiments are conducted both with and without attention.}
    \label{baseline}
\end{figure}

The baseline model for dialog generation, depicted in Figure \ref{baseline}, consists of a standard encoder-decoder framework \citep{sutskever2014sequence}, augmented with a belief tracker (obtained from the annotations of the dialog state) and a database vector. The dialog system is tasked with producing the appropriate system response, given a dialog context, an oracle belief state representation and a vector corresponding to the database output. 

The dialog context is encoded using an LSTM \citep{hochreiter1997long} sequence-to-sequence network \citep{sutskever2014sequence}. Experiments are conducted with and without an attention mechanism \citep{bahdanau2014neural}. Given the final encoder hidden state, $h_t^e$, the belief state vector, $v_{bs}$, and the database vector, $v_{db}$, Equation \ref{pol} describes how the initial decoder hidden state is obtained.

\begin{equation}
    \label{pol}
    h^d_0 = tanh \left( W_{e} h^e_t + W_{bs} v_{bs} + W_{db} v_{db} + b \right)
\end{equation}

\vspace{-0.5em}

\subsection{Neural Dialog Modules}

As seen in Figure \ref{traditional}, a traditional dialog system consists of the NLU, the DM and the NLG. The NLU maps a natural language input to a belief state representation (BS). The DM uses the belief state and some database output, to produce dialog acts (DA) for the system response. The NLG uses the dialog acts to produce a natural language response.

\begin{figure}[h]

    \includegraphics[width=1\linewidth]{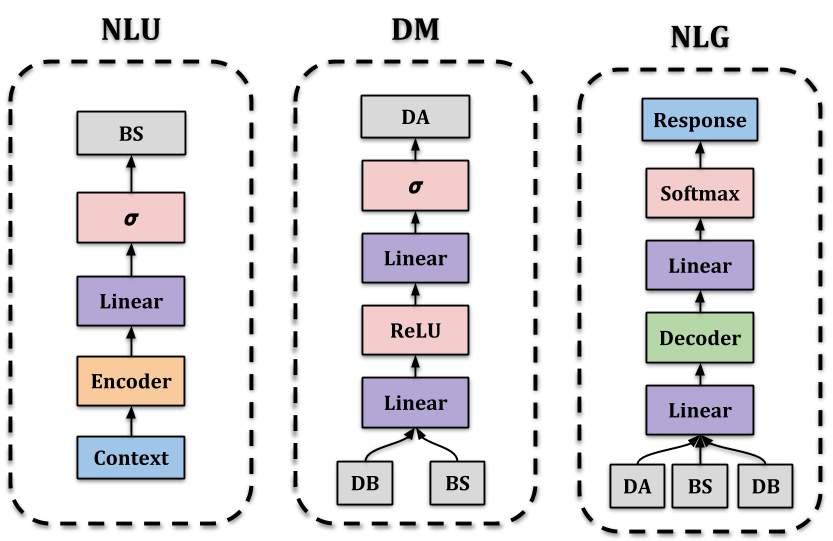}
    \caption{A visualization of the neural architectures for each of the three modules of traditional dialog systems.}
    \label{modules}
\end{figure}

A neural \textit{dialog module} is constructed for each of these three components. A visualization of these architectures is shown in Figure \ref{modules}. The NLU architecture uses an LSTM encoder to map the natural language input to a latent representation, $h_t$, which is then passed through a linear layer and a sigmoid function to obtain a multi-label prediction of the belief state. The DM architecture projects the belief state and database vector into a latent space, through the use of a linear layer with a ReLU activation, which is then passed through another linear layer and a sigmoid function to predict the dialog act vector. The neural architecture corresponding to the NLG is a conditioned language model with its initial hidden state given by a linear encoding of the dialog acts, belief state and database vectors.

The following equations define the structure of the modules, where the $gt$ subscript on an intermediate variable denotes the use of the ground-truth value:
\vspace{-0.5em}
\begin{align}
    bs &= \textbf{NLU}({\tt context}) \\
    da &= \textbf{DM}(bs_{gt}, db) \\
    {\tt response} &= \textbf{NLG}(bs_{gt}, db, da_{gt})
\end{align}

\subsection{Na\"ive Fusion}

Na\"ive Fusion (NF) is a straightforward mechanism for using the neural dialog modules for end-to-end dialog response generation.

\subsubsection{Zero-Shot Na\"ive Fusion}

During training, each dialog module is trained independently, meaning that it is given the ground truth input and supervision signal. However, during inference, the intermediate values (e.g., the dialog act vector) do not necessarily exist and the outputs of other neural modules must be used instead. For example, the DM module is trained given the ground-truth belief state as input, however during inference it must rely on the belief state predicted by the NLU module. This results in a propagation of errors, as the DM and NLG may receive imperfect input.

Zero-Shot Na\"ive Fusion combines the pre-trained neural modules at inference time. The construction of the response conditioned on the context, is described as follows:
\begin{align}
    bs &= \textbf{NLU}({\tt context}) \label{bs} \\ 
    {\tt response} &= \textbf{NLG}(bs, db, \textbf{DM}(bs, db)) \label{resp}
\end{align}

\subsubsection{Na\"ive Fusion with Fine-Tuning}

Since the forward propagation described in Equations \ref{bs} and \ref{resp} is continuous and there is no sampling procedure until the response is generated, Na\"ive Fusion can be fine-tuned for the end-to-end task of dialog generation. The pre-trained neural modules are combined as described above, and fine-tuned on the task of dialog generation using the same data and learning objective as the baseline.

\subsection{Multitask Fusion}

\begin{figure}[h]

    \includegraphics[width=1\linewidth]{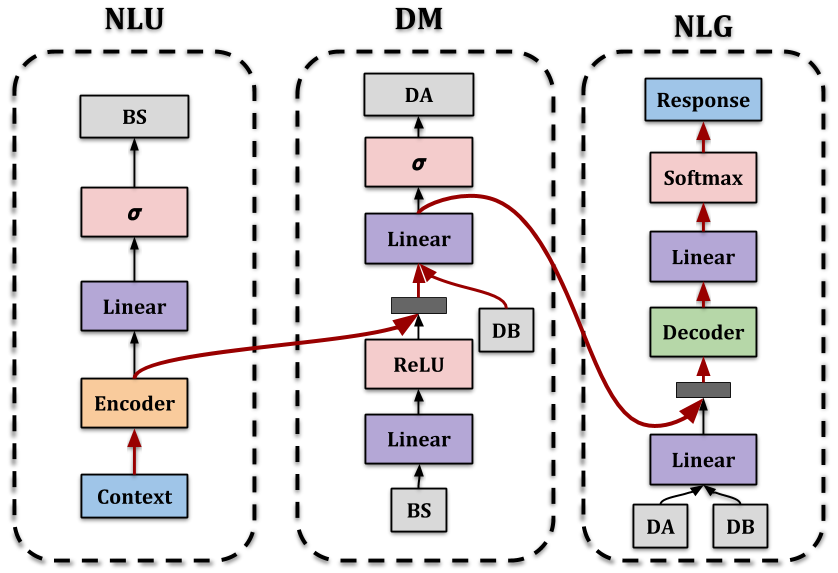}
    \caption{A depiction of Multitask Fusion, where the individual neural modules are learned simultaneously with the end-to-end task of dialog generation. The dashed boxes contain the individual components, while the red arrows depict forward propagation for the end-to-end task. The red arrows are the process used during response generation.}
    \label{multitask}
\end{figure}

Structure can be incorporated into neural architectures through the use of multi-tasking. Multitask Fusion (MF) is a method where the end-to-end generation task is learned simultaneously with the aforementioned dialog modules. The multi-tasking setup is seen in Figure \ref{multitask}.

By sharing the weights of the end-to-end architecture and each respective module, the learned representations should become stronger and more structured in nature. For example, the encoder is shared between the NLU module and the end-to-end task. As such, it will learn to both represent the information necessary for predicting the belief state vector and any additional information useful for generating the next utterance. 

\subsection{Structured Fusion Networks}

\begin{figure*}[h]

    \includegraphics[width=1\linewidth]{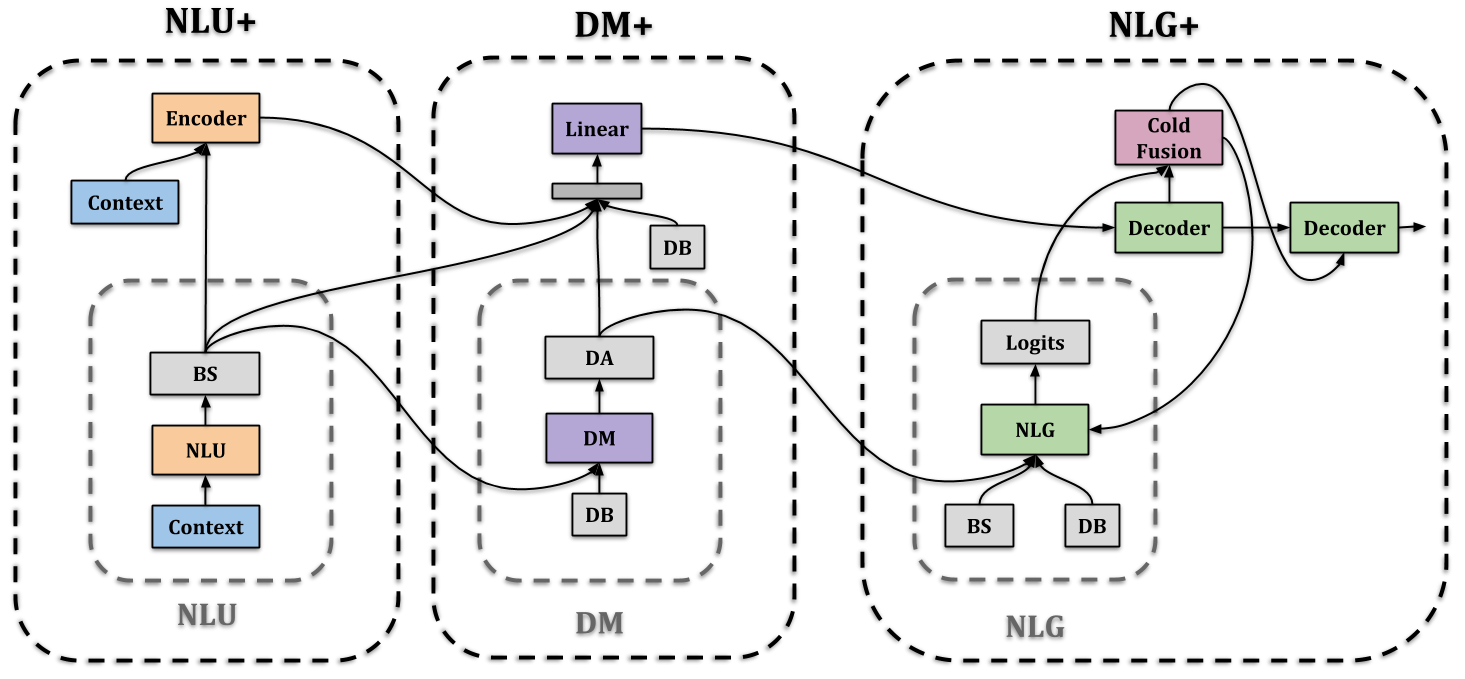}
    \caption{The Structured Fusion Network. The grey dashed boxes correspond to the pre-trained neural dialog modules. A higher-level is learned on top of the pre-trained modules, as a mechanism of enforcing structure in the end-to-end model.}
    \label{sfn}
\end{figure*}

The Structured Fusion Networks (SFNs) we propose, depicted in Figure \ref{sfn}, use the independently pre-trained neural dialog modules for the task of end-to-end dialog generation. Rather than fine-tuning or multi-tasking the independent modules, SFNs aim to learn a higher-level model on top of the neural modules to perform the task of end-to-end response generation.

The output of the NLU is concatenated at each time-step of the encoder input. The output of the DM is similarly concatenated to the input of the linear layer between the encoder and the decoder of the higher-level model. The output of the NLG, in the form of logits at a decoding time-step, is combined with the hidden state of the decoder via cold-fusion \citep{sriram2017cold}. Given the NLG output as $l_t^{NLG}$ and the higher-level decoder hidden state as $s_t$, the cold-fusion method is described as follows:

\begin{align}
    h_t^{NLG} & = DNN(l_t^{NLG}) \\
    g_t & = \sigma(W[s_t; h_t^{NLG}]+b) \\
    s_t^{CF} & = [s_t; g_t \circ h_t^{NLG}] \\
    y_t & = softmax(DNN(s_t^{CF}))
\end{align}

By pre-training the modules and using their structured outputs, the higher-level model does not have to \textit{re-learn} and \textit{re-model} the dialog structure (i.e., representing the belief state and dialog acts). Instead, it can focus on the more abstract modelling that is necessary for the task, including recognizing and encoding complex natural language input, modelling a policy, and effectively converting a latent representation into a natural language output according to the policy.

The SFN architecture may seem complicated due to the redundancy of the inputs. For example, the context is passed to the model in two places and the database vector in three places. This redundancy is necessary for two reasons. First, each of the neural modules must function independently and thus needs sufficient inputs. Second, the higher-level model should be able to function well independently. If any of the neural modules was to be removed, the SFN should be able to perform reasonably. This means that the higher-level module should not rely on any of the neural modules to capture information about the input and therefore allow the neural modules to focus only on representing the structure. For example, if the context was not passed into the higher-level encoder and instead only to the NLU module, then the NLU may no longer be able to sufficiently model the belief state and may instead have to more explicitly model the context (e.g., as a bag-of-words representation). 

Several variations of training SFNs are considered during experimentation, enumerated as follows. (1) The pre-trained neural modules are kept frozen, as a way of ensuring that the structure is not deteriorated. (2) The pre-trained neural modules are fine-tuned for the end-to-end task of response generation. This ensures that the model is able to abandon or modify certain elements of the structure if it helps with the end-to-end task. (3) The pre-trained modules are multi-tasked with the end-to-end task of response generation. This ensures that the structure is maintained and potentially strengthened while also allowing the modules to update and improve for the end-to-end task.

\begin{table*}[]
\centering
\begin{tabular}{|l|c|c|c|c|}
\hline
                    Model & BLEU   & Inform & Success  & Combined Score \\ \hline
\multicolumn{5}{|c|}{Supervised Learning}\\ \hline

Seq2Seq \citep{budzianowski2018MultiWOZ}       & 18.80   & 71.29\%   & 60.29\%    & 84.59          \\
Seq2Seq w/ Attention \citep{budzianowski2018MultiWOZ}       & 18.90   & 71.33\%   & 60.96\%    & 85.05          \\ 
Seq2Seq  (Ours)            & 20.78  & 61.40\%   & 54.50\%    & 78.73          \\
Seq2Seq w/ Attention (ours) & 20.36  & 66.50\%   & 59.50\%    & 83.36          \\         
3-layer HDSA \citep{chen2019semantically} & \textbf{23.60} & \textbf{82.90\%} & \textbf{68.90\%} & \textbf{99.50}  \\ \hline 

Na\"ive Fusion (Zero-Shot)              & 7.55  & 70.30\%   & 36.10\%    & 60.75          \\
Na\"ive Fusion (Fine-tuned Modules) & 16.39  & 66.50\%   & 59.50\%    & 83.36          \\          
Multitasking & 17.51 & 71.50\% & 57.30\% & 81.91 \\ \hline
Structured Fusion (Frozen Modules) & 17.53 & 65.80\% & 51.30\% & 76.08 \\
Structured Fusion (Fine-tuned Modules) & 18.51 & 77.30\% & 64.30\% & 89.31 \\
Structured Fusion (Multitasked Modules) & 16.70 & 80.40\% & 63.60\% & 88.71 \\
\hline
\multicolumn{5}{|c|}{Reinforcement Learning}\\ \hline
Seq2Seq + RL \citep{zhao2019rethinking} & 1.40 & 80.50\% & \textbf{79.07\%} & 81.19 \\
LiteAttnCat + RL \citep{zhao2019rethinking} & 12.80 & \textbf{82.78\%} & \textbf{79.20\%} & \textbf{93.79} \\
Structured Fusion (Frozen Modules) + RL & \textbf{16.34} &\textbf{82.70\%} & 72.10\% & \textbf{93.74} \\ \hline

\end{tabular}
\caption{Experimental results for the various models. This table compares two classes of methods: those trained with supervised learning and those trained with reinforcement learning. All bold-face results are statistically significant ($p < 0.01$).}
\label{tab:results}
\end{table*}

\section{Experiments}

\subsection{Dataset}

The dialog systems are evaluated on the MultiWOZ dataset \citep{budzianowski2018MultiWOZ}, which consists of ten thousand human-human conversations covering several domains. The MultiWOZ dataset contains conversations between a tourist and a clerk at an information center which fall into one of seven domains - attraction, hospital, police, hotel, restaurant, taxi, train. Individual conversations span one to five of the domains. Dialogs were collected using the Wizard-of-Oz framework, where one participant plays the role of an automated system.

Each dialog consists of a goal and multiple user and system utterances. Each turn is annotated with two binary vectors: a belief state vector and a dialog act vector. A single turn may have multiple positive values in both the belief state and dialog act vectors. 
The belief state and dialog act vectors are of dimensions 94 and 593, respectively.

Several metrics are used to evaluate the models. BLEU \citep{papineni2002bleu} is used to compute the word overlap between the generated output and the reference response. Two task-specific metrics, defined by \citet{budzianowski2018MultiWOZ}, Inform rate and Success rate, are also used. Inform rate measures how often the system has provided the appropriate entities to the user. Success rate measures how often the system answers all the requested attributes. Similarly to \citet{budzianowski2018MultiWOZ}, the best model is selected during validation using the combined score which is defined as $BLEU + 0.5 \times (Inform + Success)$. This combined score is also reported as an evaluation metric. 

\subsection{Experimental Settings}

The hyperparameters match those used by \citet{budzianowski2018MultiWOZ}: embedding dimension of 50, hidden dimension of 150, and a single-layer LSTM. All models are trained for 20 epochs using the Adam optimizer \cite{DBLP:journals/corr/KingmaB14}, with a learning rate of 0.005 and batch size of 64.  The norm of the gradients are clipped to 5 \cite{DBLP:journals/corr/abs-1211-5063}. Greedy decoding is used during inference.

All previous work uses the ground-truth belief state vector during training and evaluation. Therefore the experiments with the SFNs have the NLU module replaced by an "oracle NLU" which always outputs the ground-truth belief state. Table \ref{bs_ablation} in the Appendix shows experimental results which demonstrate that using only the ground-truth belief state results in the best performance.

\subsection{Reinforcement Learning}

A motivation of explicit structure is the hypothesis that it will reduce the effects of the implicit language model, and therefore mitigate degenerate output after reinforcement learning. This hypothesis is evaluated by fine-tuning the SFNs with reinforcement learning. The setup for this experiment is similar to that of \citet{zhao2019rethinking}: (1) the model produces a response conditioned on a ground-truth dialog context, (2) the success rate is evaluated for the generated response, (3) using the success rate as the reward, the policy gradient is calculated at each word, and (4) the parameters of the model are updated. A learning rate of \texttt{1e-5} is used with the Adam optimizer \citep{kingma2014adam}. 

Reinforcement learning is used to fine-tune the best performing model trained in a supervised learning setting. During this fine-tuning, the neural dialog modules (i.e., the NLU, DM and NLG) are frozen. Only the high-level model is updated during reinforcement learning. Freezing maintains the structure, while still updating the higher level components. Since the structure is maintained, it is unnecessary to alternate between supervised and reinforcement learning. 

\subsection{Results}

Experimental results in Table \ref{tab:results} show that our Structured Fusion Networks (SFNs) obtain strong results when compared to both methods trained with and without the use of reinforcement learning. Compared to previous methods trained only with supervised learning, SFNs obtain a \textbf{+4.26} point improvement over seq2seq baselines in the combined score with strong improvement in both Success and Inform rates. SFNs are outperformed by the recently published HDSA \citep{chen2019semantically} models which relies on BERT \citep{devlin2018bert} and conditioning on graph structured dialog acts. When using reinforcement learning, SFNs match the performance of LiteAttnCat \citep{zhao2019rethinking} on the combined score. Though the Inform rate is equivalent and the Success rate is lower (albeit still better than all supervised methods), the BLEU score of SFNs is much better with an improvement of \textbf{+3.54} BLEU over LiteAttnCat. 

In the reinforcement learning setting, the improved BLEU can be attributed to the explicit structure of the model. This structure enables the model to optimize for the reward (Success rate) without resulting in degenerate output \citep{lewis2017deal}. 

SFNs obtain the highest combined score when the modules are fine-tuned. This is likely because, while the structured modules serve as a strong initialization for the task of dialog generation, forcing the model to maintain the exact structure (i.e., frozen modules) limits its ability to learn. In fact, the end-to-end model may choose to ignore some elements of intermediate structure (e.g., a particular dialog act) which prove useless for the task of response generation. 

Despite strong overall performance, SFNs do show a \textbf{-2.27} BLEU drop when compared to the strongest seq2seq baseline and a \textbf{-5.09} BLEU drop compared to HDSA. Though it is difficult to ascertain the root cause of this drop, one potential reason could be that the dataset contains many social niceties and generic statements (e.g., \textit{"happy anniversary"}) which are difficult for a structured model to effectively generate (since it is not an element of the structure) while a free-form sequence-to-sequence network would not have this issue. 

To a lesser degree, multi-tasking (i.e., multitasked modules) would also prevent the model from being able to ignore some elements of the structure. However, the SFN with multitasked modules performs best on the Inform metric with a \textbf{+9.07\%} improvement over the seq2seq baselines and a \textbf{+3.10\%} over other SFN-based methods. This may be because the Inform metric measures how many of the requested attributes were answered, which benefits from a structured representation of the input. 

Zero-Shot Na\"ive Fusion performs very poorly, suggesting that the individual components have difficulty producing good results when given imperfect input. Though the NLG module performs extremely well when given the oracle dialog acts (28.97 BLEU; 106.02 combined), its performance deteriorates significantly when given the predicted dialog acts. This observation is also applicable to Structured Fusion with frozen modules. 

HDSA \citep{chen2019semantically} outperforms SFN possibly due to the use of a more sophisticated Transformer model \citep{vaswani2017attention} and BERT pre-training \citep{devlin2018bert}. A unique advantage of SFNs is that the architecture of the \textit{neural dialog modules} is flexible. The performance of HDSA could potentially be integrated with SFNs by using the HDSA model as the NLG module of an SFN. This is left for future work, as the HDSA model was released while this paper was already in review.

These strong performance gains reaffirm the hypothesis that adding explicit structure to neural dialog systems results in improved modelling ability particularly with respect to dialog policy as we see in the increase in Inform and in Success. The results with reinforcement learning suggest that the explicit structure allows \textit{controlled fine-tuning} of the models, which prevents divergent behavior and degenerate output.

\subsection{Human Evaluation}

To supplement the results in Table \ref{tab:results}, human evaluation was used to compare seq2seq, SFN, SFN fine-tuned with reinforcement learning, and the ground-truth human response. Workers on Amazon Mechanical Turk (AMT) were asked to read the context, and score the \textit{appropriateness} of each response on a Likert scale (1-5). One hundred context-response pairs were labeled by three workers each. The results shown in Table \ref{tab:human} demonstrate that SFNs with RL outperform the other methods in terms of human judgment. These results indicate that in addition to improving on automated metrics, SFNs result in user-favored responses.

\begin{table}[h]
\begin{tabular}{|c|c|c|c|}
\hline
Model             & Avg. Rating  & $\geq$ 4  & $\geq$ 5 \\
\hline
Seq2Seq           & 3.00 & 40.21\% & 9.61\%  \\
SFN               & 3.02 &\textbf{ 44.84\% }& 11.03\%  \\
SFN + RL          & \textbf{3.12} &\textbf{ 44.84\%} & \textbf{16.01\%}  \\
\hline
Human             & 3.76 & 59.79\% & 34.88\%  \\
\hline
\end{tabular}
\caption{Results of human evaluation experiments. The $\geq$ 4 and $\geq$ 5 columns indicate the percentage of system outputs which obtained a greater than 4 and 5 rating, respectively.}
\label{tab:human}
\end{table}

\section{Analysis}

\label{sec:analysis}

\subsection{Limited Data}

Structured Fusion Networks (SFNs) should outperform sequence-to-sequence (seq2seq) networks in reduced data scenarios due to the explicit structure. While a baseline method would require large amounts of data to learn to infer structure, SFNs do this explicitly. 

The performance of seq2seq and SFNs are determined, when training on 1\%, 5\%, 10\% and 25\% of the training data (total size of $\sim$ 55,000 utterances). The supervised-learning variant of SFNs with fine-tuned modules is used. The pre-training of the modules and fine-tuning of the full model is done on the same data split. The full data is used during validation and testing.

\begin{figure}[h]
    \vspace{-1.0em}
    \centering
    \begin{subfigure}[b]{0.85\columnwidth}
    \includegraphics[width=1\linewidth]{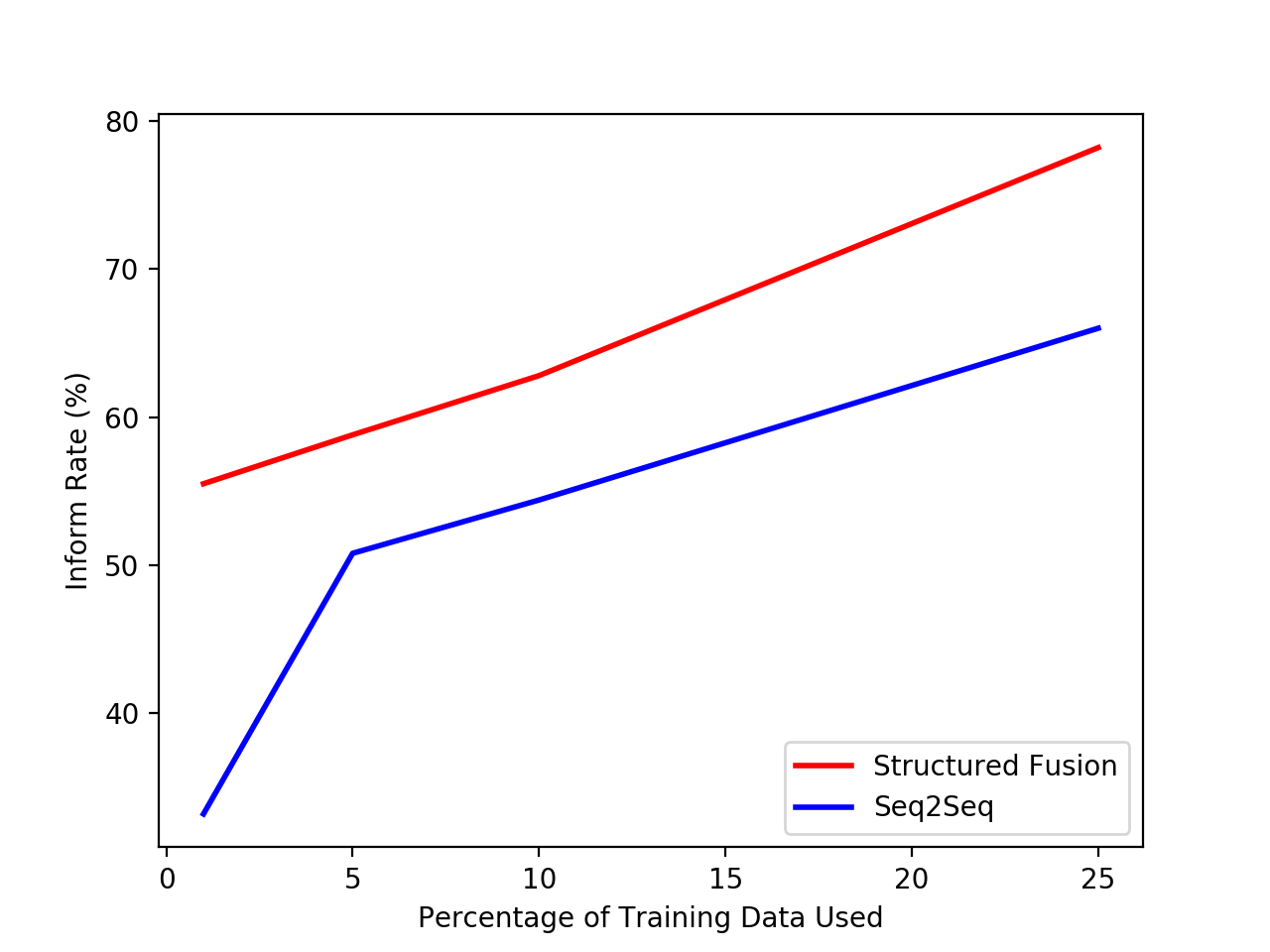}
    \caption{}
    \label{fig:Ng1} 
    \end{subfigure}
    \vspace{-0.5em}
    \begin{subfigure}[b]{0.85\columnwidth}
    \includegraphics[width=1\linewidth]{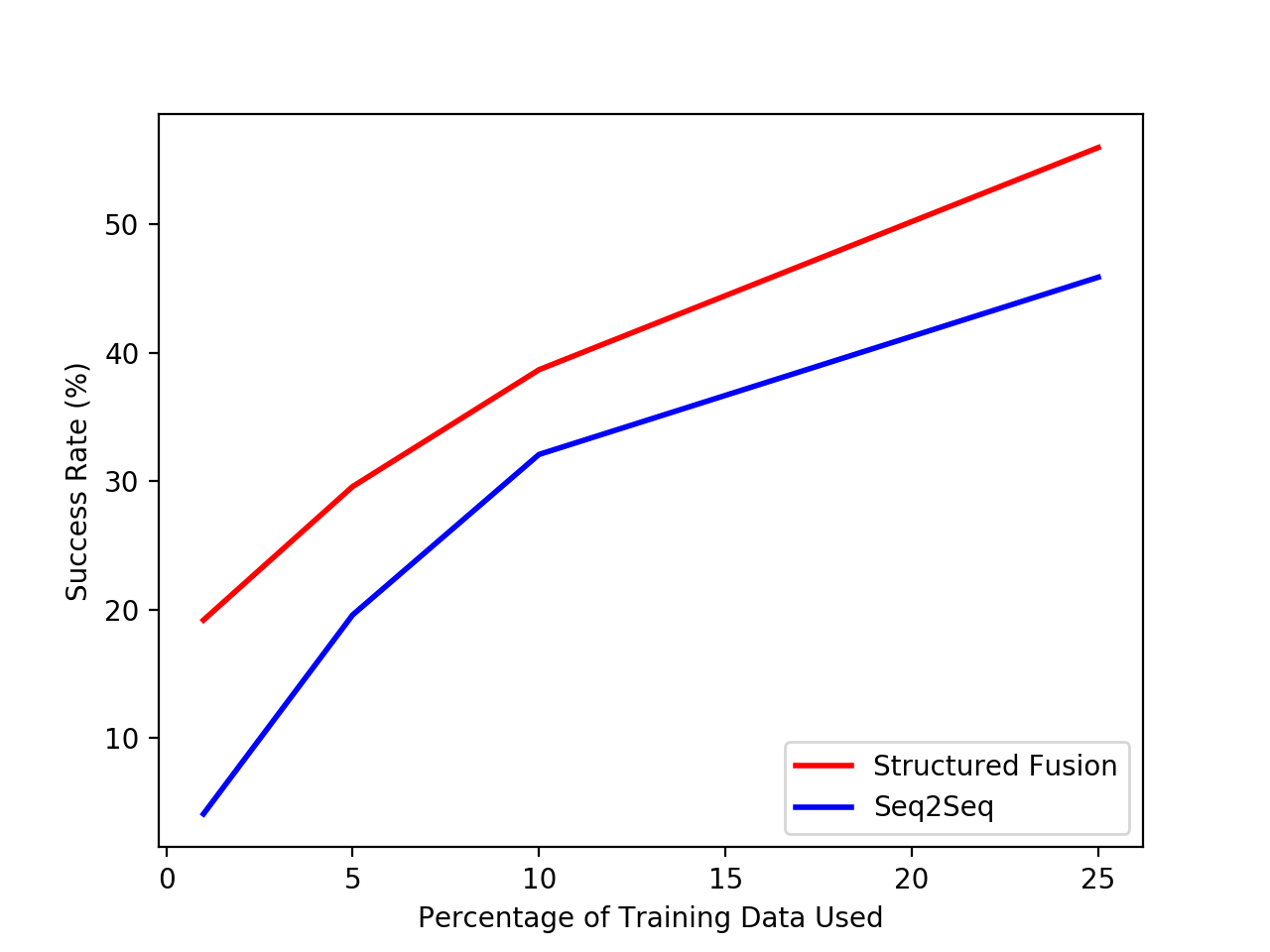}
    \caption{}
    \label{fig:Ng2}
    \end{subfigure}
    \caption{Variation of Inform (a) and Success (b) rate at different amounts of training data.}
    \label{fig:low-resource}
    \vspace{-0.5em}
\end{figure}

The results in Figure \ref{fig:low-resource} show the Inform and Success rates for different amounts of training data. SFNs significantly outperform the seq2seq model in low-data scenarios. Notably, improvement is considerably higher in the most extreme low-data scenario, when only 1\% of the training data ($\sim$ 550 dialogs) is used. As the amount of training data increases, the gap between the two models stabilizes. The effectiveness at extreme low-data scenarios reaffirms the hypothesis that explicit structure makes SFNs less data-hungry than sequence-to-sequence networks.

\subsection{Domain Generalizability}

The explicit structure of SFNs should facilitate effective domain generalizability. A domain transfer experiment was constructed to evaluate the comparative ability of seq2seq and SFNs. The models were both trained on a reduced dataset that largely consists of out-of-domain examples and evaluated on in-domain examples. Specifically, 2000 out-of-domain training examples and only 50 in-domain training examples were used. The restaurant domain of MultiWOZ was selected as in-domain. 



\begin{table}[h]
\begin{tabular}{|c|c|c|c|}
\hline
Model             & BLEU  & Inform  & Success \\
\hline
Seq2Seq           & 10.22 & 35.65\% & 1.30\%  \\
Structured Fusion & 7.44  & \textbf{47.17}\% & \textbf{2.17}\% \\
\hline
\end{tabular}
\caption{Results of the domain transfer experiment comparing sequence-to-sequence and Structured Fusion Networks. All bold-face results are statistically significant ($p < 0.01$).}
\label{tab:domain-transfer}
\vspace{-0.5em}
\end{table}

The results, seen on Table \ref{tab:domain-transfer}, show that SFNs perform significantly better on both the Inform (\textbf{+11.52\%}) and Success rate. Although SFNs have a slightly higher Success rate, both models perform poorly. This is expected since the models would be unable to answer all the requested attributes when they have seen little domain data -- their language model would not be tuned to the in-domain task. The \textbf{-2.78} BLEU reduction roughly matches the BLEU difference observed on the main task, therefore it is not an issue specific to domain transfer.

\section{Conclusions and Future Work}

This paper presents several methods of incorporating explicit structure into end-to-end neural models of dialog. We created Structured Fusion Networks, comprised of pre-trained \textit{dialog modules} and a higher-level end-to-end network, which obtain strong results on the MultiWOZ dataset both with and without the use of reinforcement learning. SFNs are further shown to be robust to divergence during reinforcement learning, effective in low data scenarios and better than sequence-to-sequence on the task of domain transfer. 
 
For future research, the explicit structure of SFNs has been shown to have multi-faceted benefits; another potential benefit may be interpretability. It would be interesting to investigate the use of SFNs as more interpretable models of dialog. While domain generalizability has been demonstrated, it would be useful to further explore the nature of generalizability (e.g., task transfer, language style transfer). Another potential avenue of research is whether the explicit structure of SFNs could potentially allow swapping the dialog modules without any fine-tuning. Structured Fusion Networks highlight the effectiveness of using explicit structure in end-to-end neural networks, suggesting that exploring alternate means of incorporating structure would be a promising direction for future work.

\bibliography{acl2018}
\bibliographystyle{acl_natbib}
\clearpage
\appendix
\section{Belief State Ablation Study }

All previous research working on dialog generation for the MultiWOZ dataset uses the ground-truth belief state vector during training and evaluation. Therefore for fair comparability, the SFN experiments in our paper had the NLU module replaced by an "oracle NLU" which always outputs the ground-truth belief state.

An ablation experiment was performed to ascertain whether providing \textit{only} the ground-truth belief state was the optimal solution. Several methods of combining the ground-truth belief state with the pre-trained NLU module were explored. These methods are enumerated as follows:

\begin{enumerate}
    \item[(1)] \textbf{Ground-Truth Only:} The setting used in the primary experiments, shown in Table 1 of the main paper. Only the ground-truth belief state vector is used.
    \item[(2)] \textbf{Predicted Only:} Only the belief state predicted by the pre-trained NLU module is used. 
    \item[(3)] \textbf{Sum:} The predicted and ground-truth belief states are summed, before being used by all upper layers.
    \item[(4)] \textbf{Linear:} The predicted and ground-truth belief states area concatenated and passed through a linear layer. 
\end{enumerate}

These experiments are performed using the best model, Structured Fusion Networks with fine-tuned modules. The results are shown in Table \ref{bs_ablation}.

\begin{table}[h]
\centering
\begin{tabular}{|c|c|c|c|c|}
\hline
Model             & BLEU  & Inform  & Success & Comb. \\
\hline
GT               & \textbf{18.51}&\textbf{77.30\%} & \textbf{64.30\%} & \textbf{89.31} \\
Pred             & 16.88 & 73.80\% & 58.60\% & 83.04  \\
Sum              & 15.93 & 72.90\% & 60.80\% & 82.78  \\
Linear           & 15.42 & 66.80\% & 54.80\% & 76.22  \\
\hline
\end{tabular}
\caption{Results of the domain transfer experiment comparing sequence-to-sequence and Structured Fusion Networks. All bold-face results are statistically significant ($p < 0.01$).}
\label{bs_ablation}
\vspace{-0.5em}
\end{table}

It is observed that adding the pre-trained NLU does not provide any additional performance benefit, when the ground-truth belief state is already provided. As such, combinations of the ground-truth and predicted belief state actually perform worse than either of the methods independently because of (1) additional parameters to be learned, especially in the case of the \textit{Linear} method, and (2) a conflicting trade-off between fine-tuning a learned NLU module and using the ground-truth belief state. 

\section{Qualitative Examples}

Table \ref{examples} shows several examples of dialogs from the test set of MultiWOZ, along with the produced response from three different models: sequence-to-sequence networks, Structured Fusion Networks, and Structured Fusion Networks fine-tuned with reinforcement learning. These examples serve to provide insight into the respective strengths and weaknesses of the different models. A few noteworthy observations from the four examples are enumerated below:

\begin{enumerate}
    \item[(1)] SFN fine-tuned with RL \textbf{consistently provides more attribute information.} It provides at least one attribute in every example response, for a total of 14 total attributes across the four examples. This, along with the high Success score of this model, is a consequence of the reinforcement learning fine-tuning which directly optimizes the Success score and rewards the model for providing the correct attributes. 
    \item[(2)] Seq2Seq \textbf{produces more generic responses}. In the second and fourth examples, the outputs produced by Seq2Seq are generic and are  unrelated to the context. This suggests that the Seq2Seq model has an overwhelming implicit language model, which produces generic and dull responses. On the other hand, the explicit structure in SFN mitigates the effect of the implicit language model by relying on an \textit{explicit} language model (in the form of the NLG).
    \item[(3)] Seq2Seq \textbf{requests attributes which the user has already provided.} In the first and third example, the Seq2Seq output is requesting parameters that the user explicitly provided. This highlights the lack of structured representations in the Seq2Seq model. SFN which explicitly models the structure of the input effectively understands and captures the information provided in the context. 
\end{enumerate}

\begin{table*}[]
    \centering
    \begin{tabular}{l}
    \hline \hline
    \multicolumn{1}{c}{\textbf{Example 1}}\\
    \hline \hline
\textbf{Dialog Context:}\\
\texttt{USER:} hello . i am looking for a [value\_pricerange] place to eat in the [value\_area] . can you help me ? \\
\hline 
\textbf{Sequence-to-Sequence Response:} \\
i have [value\_count] options for you . do you have a preference on price range ?\\
\hline 
\textbf{Structured Fusion Network Response:} \\
\. [restaurant\_name] is  a [value\_pricerange] -ly priced [value\_food] restaurant in the [value\_area] . would \\
you like to make a reservation ?\\
\hline 
\textbf{Structured Fusion Network w/ Reinforcement Learning Response:} \\
\.[restaurant\_name] is a [value\_pricerange] [value\_food] restaurant in the [value\_area] . it is located\\ at [restaurant\_address] . the postcode is [restaurant\_postcode] . the phone number is [restaurant\_phone] \\

\hline \hline
    \multicolumn{1}{c}{\textbf{Example 2}}\\
    \hline \hline
\textbf{Dialog Context:}\\
\texttt{...} \\
\texttt{USER:} i want to book this hotel for the same group of people and [value\_count] nights starting \\from [value\_day] . i'll need a reference number please . \\
\texttt{SYSTEM:} your reservation is confirmed and the reference number is \#\#\\
\texttt{USER:} i am sorry ? were you able to book the express by holiday inn for [value\_count] people ? \\
\hline 
\textbf{Sequence-to-Sequence Response:} \\
you are welcome . have a great day !\\
\hline 
\textbf{Structured Fusion Network Response:} \\
you are all set . your reference number is [hotel\_reference] . can i help you with anything else ? \\
\hline 
\textbf{Structured Fusion Network w/ Reinforcement Learning Response:} \\
your booking was successful . the reference number is [hotel\_reference] . \\

\hline \hline
    \multicolumn{1}{c}{\textbf{Example 3}}\\
    \hline \hline
\textbf{Dialog Context:}\\
\texttt{USER:} can you see if there are any entertainment attractions in the [value\_area] of town ? \\
\hline 
\textbf{Sequence-to-Sequence Response:} \\
there are [value\_count] museums in the [value\_area] . do you have a preference for area ?\\
\hline 
\textbf{Structured Fusion Network Response:} \\
there are [value\_count] museums in the [value\_area] . [attraction\_name] is a museum and the [attraction\_name] \\ is a [value\_count] . \\
\hline 
\textbf{Structured Fusion Network w/ Reinforcement Learning Response:} \\
\. [attraction\_name] is a museum in the [value\_area] and the [attraction\_name] is located at [attraction\_address] . \\ the postcode is [attraction\_postcode] . \\

\hline \hline
    \multicolumn{1}{c}{\textbf{Example 4}}\\
    \hline \hline
\textbf{Dialog Context:}\\
\texttt{USER:} i need a taxi . i'll be departing from [restaurant\_name] . \\
\texttt{SYSTEM:} i can help you with that . when do you need to leave ? \\
\texttt{USER:} i would like to leave after [value\_time] please .  \\
\texttt{SYSTEM:} where will you be going ? \\
\texttt{USER:} i'll be going to [restaurant\_name] . \\
\hline 
\textbf{Sequence-to-Sequence Response:} \\
you are welcome . have a good time !\\
\hline 
\textbf{Structured Fusion Network Response:} \\
you are welcome . have a great day ! \\
\hline 
\textbf{Structured Fusion Network w/ Reinforcement Learning Response:} \\
i have booked for you a [taxi\_type] , the contact number is [taxi\_phone] . is there anything else i can\\ help you with ? \\

\hline 

\hline
\end{tabular}
\caption{Four examples of dialog contexts from the dataset, and the responses generated by three different models: Seq2Seq, SFN, and SFN with RL.}
\label{examples}
\end{table*}

\end{document}